# Band-to-Band Tunneling based Ultra-Energy Efficient Silicon Neuron

Tanmay Chavan[1]*, Sangya Dutta[1], Nihar R. Mohapatra[2], and Udayan Ganguly[1]†

[1]Department of Electrical Engineering, IIT Bombay, Mumbai 400076, India
[2]Department of Electrical Engineering, IIT Gandhinagar, Gandhinagar 382355, India

*tanmay.chavan@iitb.ac.in; †udayan@ee.iitb.ac.in

## Abstract

The human brain comprises about a hundred billion neurons connected through quadrillion synapses. Spiking Neural Networks (SNNs) take inspiration from the brain to model complex cognitive and learning tasks. Neuromorphic engineering implements SNNs in hardware, aspiring to mimic the brain at scale (i.e., 100 billion neurons) with biological area and energy efficiency. The design of ultra-energy efficient and compact neurons is essential for the large-scale implementation of SNNs in hardware. In this work, we have experimentally demonstrated a Partially Depleted (PD) Silicon-On-Insulator (SOI) MOSFET based Leaky-Integrate & Fire (LIF) neuron where energy-and area-efficiency is enabled by two elements of design - first tunneling based operation and second compact sub-threshold SOI control circuit design. Band-to-Band Tunneling (BTBT) induced hole storage in the body is used for the "*Integrate*" function of the neuron. A compact control circuit "*Fires*" a spike when the body potential exceeds the firing threshold. The neuron then "*Resets*" by removing the stored holes from the body contact of the device. Additionally, the control circuit provides "*Leakiness*" in the neuron which is an essential property of biological neurons. The proposed neuron provides 10× higher area efficiency compared to CMOS design with equivalent energy/spike. Alternatively, it has $10^4$ × higher energy efficiency at area-equivalent neuron technologies. Biologically comparable energy- and area-efficiency along with CMOS compatibility make the proposed device attractive for large-scale hardware implementation of SNNs.

## Background

Neuromorphic computing aims to emulate the information-processing paradigm used by the human brain to solve complex cognitive tasks like learning and recognition. Neuromorphic architectures generally implement spiking neural networks (SNNs), third generation neural networks, which offer energy- and area-efficient realization of the human brain[1]. The human brain consists of about $10^{11}$ neurons[2]. Thus, neurons are the integral components of any neuromorphic system. Design of compact, energy-efficient neurons with high manufacturability is critical for the large-scale realization of SNNs in hardware. In addition to conventional silicon-based CMOS circuits, various non-silicon material based devices have also been explored as artificial neurons. Traditional silicon-based CMOS circuit[2-16] implementations occupy a large area. Other non-silicon based neurons such as IMT[17,18], Neuristor[19], PCM[20], and PCMO[21] have new materials requirements at various levels of maturity, which add process technology complexity and cost. Previously, our group has experimentally demonstrated an impact-ionization based LIF neuron using High Volume Manufactured (HVM) Silicon-on-Insulator (SOI) technology with excellent energy- and area-efficiency[13]. Impact ionization is a high drain-current process where a small fraction ($< 0.1\%$) of the drain current generates electron-hole pairs (EHP) for floating body effect. Alternatively, a tunneling current has 100% EHP generation to enable energy efficiency. Thus, we had proposed a tunneling based neuron by a simulations study – sans experimental demonstration of tunneling based LIF behavior or energy efficient sub-threshold circuits design[22].

In this work, we demonstrate LIF behavior experimentally in a 32nm PDSOI CMOS technology and design a sub-threshold reset circuit to enable energy efficiency. Integration of Band-to-Band Tunneling current from the drain into the body to charge up the floating body coupled with leakage from the source and body terminal enables the leaky integration. Sub-threshold operation of the device enables ultra-energy efficient operation. The modulation of spiking frequency as a function of the input drain voltage amplitude is experimentally demonstrated to show LIF behavior. Finally, we design the reset circuit using subthreshold operation, validate the operation and evaluate its energy and area efficiency. Thus, we demonstrate a compact, energy efficient, and highly manufacturable LIF neuron on a 32nm PDSOI CMOS technology.

- Spiking Neural Network & LIF Neuron Model

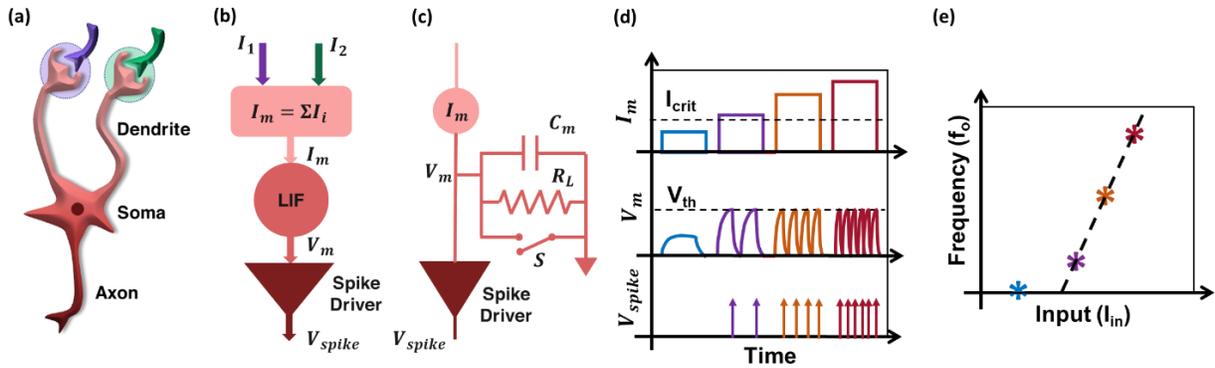

Fig. 1: (a) Biological neuron structure consists of the dendrites, soma, and the axon which are related to (b) algorithmic neuronal functionality, (c) The equivalent circuit implementation of the neuron in an SNN, (d) Transient response for LIF neuron for different inputs is shown. For input ($I_m < I_{crit}$), the neuron never spikes since $V_m$ never exceeds $V_{th}$. The frequency is zero in this case. However, for $I_m \geq I_{crit}$, neuron fires when $V_m$ exceeds $V_{th}$ and the frequency of firing increases with an increase in $I_m$, (e) Typical output frequency ($f_o$) vs. input current characteristics of an LIF neuron which is to be mimicked artificially.

The structure of a typical biological neuron is shown in Fig. 1(a). It uses dendrites to collect currents ($I_i$) and sums them ($I_m = \Sigma I_i$) from various synapses. The input current ($I_m$) is used by the Soma to perform the LIF function. Finally, every instant the neuron triggers a firing event, a spike (pulse) is sent out through the axon to the other neurons by a spike driver. The algorithmic representation of the computations performed by the biological neuron[23] is shown in Fig. 1(b). The circuit equivalent model for the LIF neuron consists of a parallel connected leak resistance ($R_L$) and membrane capacitor ($C_m$) as shown in Fig. 1(c). A current source $I_m$ models the net synaptic input current which charges up the capacitor and produces a potential $V_m$ equivalent to the neuron membrane potential. When the potential exceeds the threshold ($V_m \geq V_{th}$), the neuron is reset to the resting potential ($E_L$) by discharging the capacitor through the voltage-controlled switch (S), akin to a biological neuron. A separate Spike Driver is used which issues a spike each time $V_m$ exceeds $V_{th}$. The governing differential equation for the LIF model is given in Equation (1).

$$C_m \frac{dV_m}{dt} = -\frac{(V_m - E_L)}{R_L} + I_m \qquad (1)$$

The transient response of the LIF neuron is shown in Fig. 1(d). At low input current ($I_m$), $V_m$ never exceeds the threshold $V_{th}$ - which produces no spikes. However, when $I_m$ is high ($I_m > I_{crit}$), the increase in input current reduces the charge up time to $V_{th}$. This leads to the increase in the output frequency ($f_o$) with an increase in input $I_m$ as shown in Fig. 1(e).

## Device Structure, Concept, and Operation

In this work, LIF neuron functionality is demonstrated using a simple PDSOI MOSFET with body contact. The device structure and circuit schematic are shown in Fig. 2(a). The neuron can operate with voltage or current input[13]. For the sake of simplicity, a voltage-based operation is explained throughout the text. The current-based operation is shown in Supplementary Information 1. The body of the device can be thought of as the soma of the biological neuron and the body potential as the membrane potential. The gate terminal is grounded and is used as the reference terminal. $V_{SG}$ of the device is set to a small positive value ($V_{SG} = 0.4\ V$) to bias the device in deep sub-threshold. The input is provided at the drain terminal ($V_{DG} = V_{input}$) and the voltage output is taken from the body terminal ($V_{BG} = V_{body}$). The leaky-integration and reset function of the neuron is explained with the help of band diagrams as shown in Fig. 2(b) and (c) respectively. To enable *leaky-integration* (Fig. 2(b)), a large bias voltage is applied at the drain terminal ($V_{DG} \sim 1V$), this causes the minority electrons in the body to tunnel into the drain, leaving behind holes resulting in current $J_{in}$ that is integrated to charge up the floating body. A small fraction of the holes leaks out over the body-source junction and through the non-ideal switch (SW), which results in leak current $J_{leak}$. Body charging takes place due to net current $J_{in} - J_{leak}$. Thus, for constant applied bias, an increasing number of holes get stored in the body of the device. The hole storage saturates when a steady-state ($J_{in} \approx J_{leak}$) is reached between the incoming hole current ($J_{in}$) and the outgoing leakage current ($J_{leak}$). The increase in stored holes in the body leads to an increase in the body potential of the device over time. Once the body potential exceeds a pre-defined threshold voltage ($V_{th}$) the neuron is *reset* by removing the stored holes. This is achieved by simply closing the switch (SW) to ground the body terminal as shown in Fig. 2(c). Thus, a completely sub-threshold operation of the transistor is designed to enable ultra-low current operation.

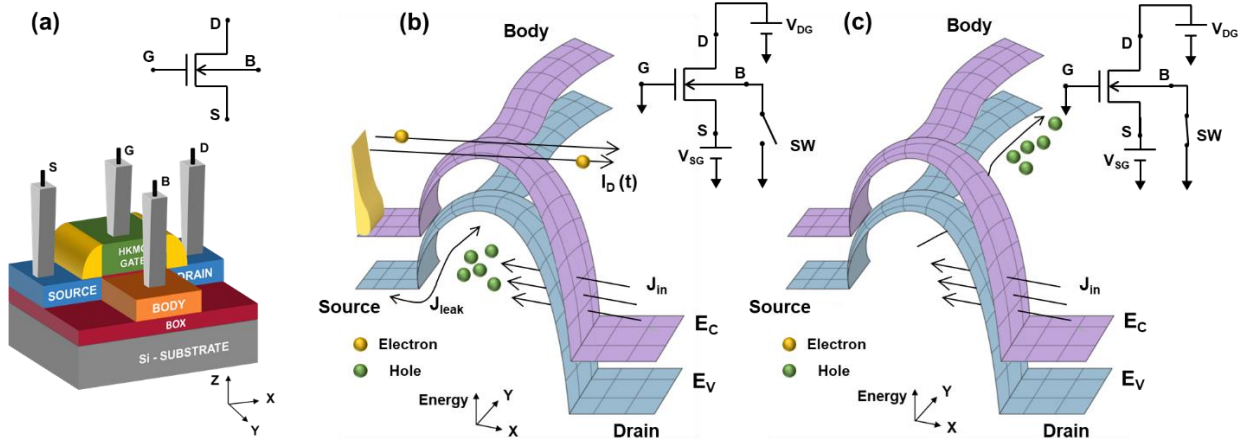

Fig. 2: (a) 3D Device schematic of the body contacted PDSOI MOSFET with circuit schematic, (b) band diagram in the x-y plane in the channel region, large drain bias causing tunneling induced hole storage (integration) when the body contact is left floating. The stored holes increase the potential of the floating body terminal akin to membrane potential. There is inherent *leakiness* due to hole leakage over the source-body barrier and through the non-ideal switch (SW), (c) band diagram in the x-y plane in the channel region showing how the neuron can be *reset* by removing the stored holes in the body by closing the switch (SW) to ground the body terminal.

For ultra-low current LIF neuron operation, an external fire and reset circuitry (shown in Fig. 3) is designed to operate in sub-threshold to enable ultra-low energy operation. The circuit takes inspiration from the integrate and fire neuron circuit proposed by Carver Mead[24]. It detects when the body potential exceeds $V_{th}$, produces a spike and resets the body (equivalent to membrane) potential by draining the body charge. The fire and reset circuit consist of two CMOS inverters (INV1 and INV2) connected in series and a MOSFET M2, which behaves as a switch. The output spikes of the neuron are obtained at the output of INV2 ($V_{spike}$), this output also drives the gate of M2. Initially, the body potential is below $V_{th}$ (defined by the switching threshold of INV1) thus, the output of INV1 ($V_X$) is high and $V_{spike}$ is low. When the input voltage is applied at the drain terminal, the body potential begins to rise due to BTBT current which is integrated to charge up the body. When the body potential crosses $V_{th}$, $V_X$ becomes low. This causes $V_{spike}$ to become high and the neuron is said to *fire*. The high voltage at $V_{spike}$ makes the switch M2 highly conducting thereby removing all the holes stored in the body of the device (M1) and the neuron resets. The $V_{GS}$ of M2 set by external leak voltage ($V_{leak}$) controls the leakage current through the body contact. All circuit elements are designed to operate in sub-threshold to enable energy efficient operation.

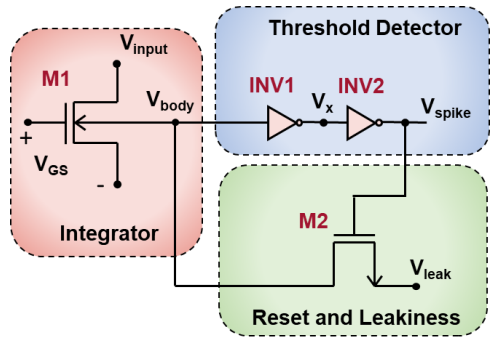

Fig. 3: The complete LIF neuron circuit, M1 is the body contact PDSOI device which performs the leaky-integration with input ($V_{input}$) applied at the drain terminal. $V_{SG}$ is set to a small positive value to bias M1 in subthreshold. INV1, INV2 & M2 perform the fire and reset of the neuron, output spikes are obtained at $V_{spike}$ and the amount of leakiness is controlled by $V_{leak}$.

**Experimental Validation**

The DC transfer characteristics ($I_D$-$V_{SG}$ with gate grounded) of the PDSOI MOSFET (W = 1$\mu m$ and L = 40 nm) in the body grounded configuration is shown in Fig. 4(a). At low drain-gate bias ($V_{DG}$) of 0.5 V there is minimal drain-body tunneling thus the drain current matches with the source current. However, when $V_{DG}$ is increased to 1 V there is significant band-to-band tunneling (BTBT) of minority electrons from the body into the drain. This leads to a significant drain current in the subthreshold region as compared to the source current. The presence of body-drain tunneling is further validated by experimentally measuring the drain, source, body and gate current as a function of drain bias ($V_{SG}$ = 0.4V, gate and body grounded) as shown in Fig. 4(b). For $V_{DG}$ greater than 0.5 V significant tunneling current is observed between the drain and the body. Also, the relatively negligible leakage current in the gate terminal rules out the possibility of gate tunneling at high $V_{DG}$. The sub-threshold BTBT current, which is much smaller than on-state current by 4 orders, is utilized for energy efficient operation.

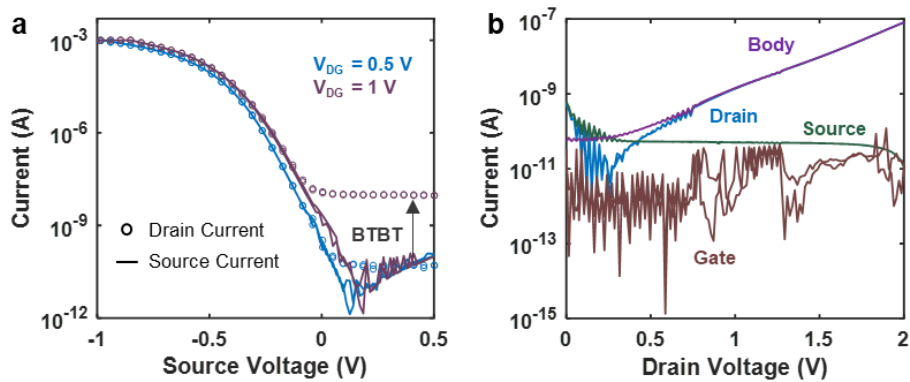

Fig. 4: (a) Experimental DC $I_D - V_{SG}$ characteristics of the device with the body grounded shows significant drain-body tunneling, (b) drain, source, gate and body current of the device as a function of drain bias at $V_{SG} = 0.4\ V$. The gate and body terminals are grounded for this measurement. For drain voltage greater than 0.5V the drain current ≈ body current. This current will lead to body charging if the body contact is left floating.

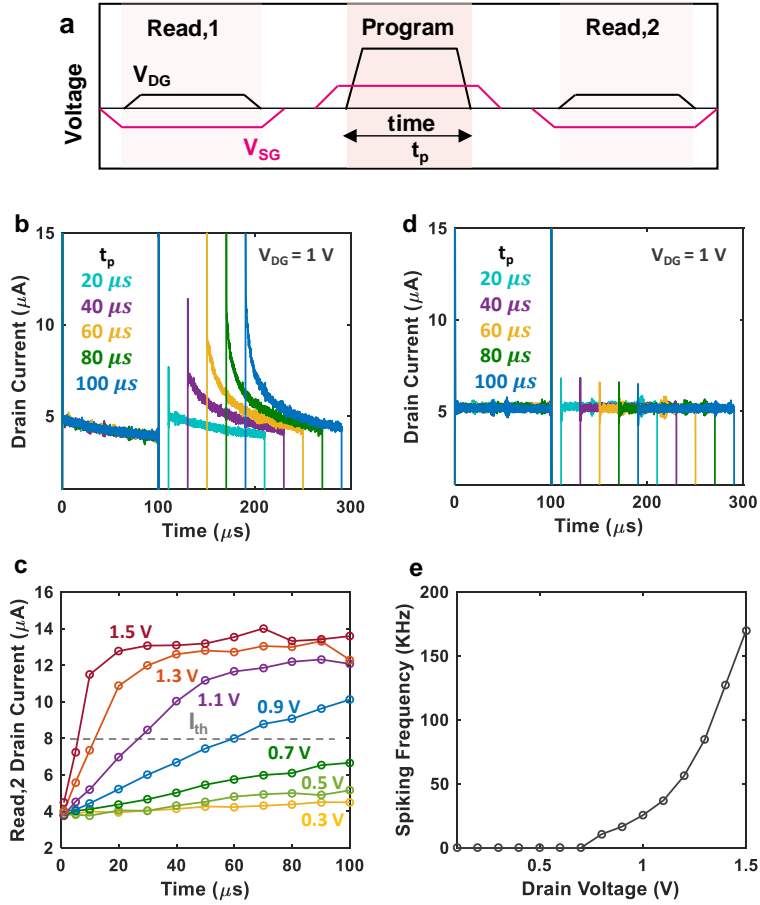

Fig. 5: (a) Read(R1)-Program(P)-Read(R2) scheme is used where the program time ($t_p$) is varied. (b) For the floating body configuration, the R1 drain current is identical and low for all case. After programming, the R2 drain current is high (due to stored holes during programming), followed by a decay toward R1 read current values (hole loss during read). An increase in program time ($t_p$) produces an increase in R2 drain current, indicating the accumulation of holes (integration) over time. All current spikes indicate displacement currents due to voltage switching, (c) High R2 drain current ($I_{D0,R2}$) increases with $t_p$. Increase in $V_{DG}$ shows an increase in the rate of integration. Comparing the current with a reference current ($I_{Th}$) produces the time to spike or the inverse of spiking frequency ($1/f_0$), (d) Same scheme applied for body ground configuration shows no increase in R2 current implying no hole storage due to efficient hole leak-out through body contact, (e) Spiking frequency ($f_0$) vs $V_{DG}$ shows typical LIF neuron behavior.

While low current and voltage can be measured *slowly* in DC with SMUs or measured *fast* by on-chip inverter gates with tera-ohm impedance and small capacitance, *off-chip fast* measurement of floating body voltage is challenging as Oscilloscopes have a low maximum impedance (1 mega-ohm) compared to on-chip inverters. To circumvent this problem, a high drain current read is used before and after the program to enable fast time-resolved measurement. Hence a Read (R1) – Program (P) – Read (R2) scheme is used to demonstrate the integration and reset functionality of neuron as shown in Fig. 5(a). In the R1 phase, the device is biased in saturation ($V_{SG}$ = -0.35 V and $V_{DG}$ = 0.1 V) and the reference drain current ($I_{D,R1}$) is measured. In the P phase, a large $V_{DG}$ (1 V) and a small positive $V_{SG}$ (0.4 V) is applied to enable leaky-integration. In R2 phase the device is biased identical to the R1 phase and the drain current ($I_{D,R2}$) is measured. To demonstrate integration, we measure the device in body contact floating condition as shown in Fig. 5(b). The drain currents show displacement current spikes when voltage switching occurs during the transition to R1, P and R2 conditions. Initially, the R1 drain current is low (~$5\mu A$). The P drain current (in subthreshold) is orders of magnitude lower and cannot be observed in the scale. The R2 drain current is initially higher than R1 (due to the holes stored in the body during P) and then decays slowly towards R1 drain current levels (due to body source leakage). The initial R2 drain current ($I_{D0,R2}$) is a measure of holes stored during P. A gradual increase in $I_{D0,R2}$ followed by saturation with an increase in program time ($t_p$) is observed. This indicates the gradual accumulation of holes (*integration*) in the floating body with time. The $I_{D0,R2}$ current is then sampled and plotted as a function of $V_{DG}$ and $t_p$ in Fig. 5(c). The rate of integration increases by increasing $V_{DG}$. To demonstrate reset, we measure with body contact shorted to ground to discourage charge storage. When the same scheme is applied to the device in body grounded configuration no increase in $I_{D,R2}$ is seen (Fig. 5(d)). In this configuration, the holes flow out of the body contact thus there is no hole storage in this configuration. Hence reset can be performed in the neuron by grounding the body contact. The comparison of $I_{D0,R2}$ (Fig. 5(c)) to a threshold ($I_{Th}$) produces a spike time which is inverse of the spike frequency ($1/f_o$). The spike frequency ($f_0$) vs input voltage ($V_{DG}$) curve shows typical LIF characteristics (Fig. 5(e)).

**Performance & Benchmarking**

The area and energy performance of the circuit is evaluated by implementing the neuron circuit in Cadence Virtuoso using GLOBALFOUNDRIES GFUS7SW technology. Detailed information about layout, transient simulations, and energy/spike for the neuron are given in Supplementary Information 1. A behavioral model is developed for the neuron which is incorporated in an SNN to solve Fischer's Iris classification problem. The network demonstrates state-of-the-art learning performance with recognition accuracy of 96% (Supplementary Information 2). The hardware translation of this SNN is explored extensively in the literature[25–27]. The area and energy/spike of the neuron is benchmarked with literature in Table 1. The conventional CMOS based neurons occupy large chip area and consume considerable energy/spike[5,9,13–16]. Recently an energy efficient CMOS neuron is demonstrated[6] but has poor area efficiency. Active efforts are being made to use novel materials like PCM[20] and PCMO[21] to improve the area and energy efficiency, but in their present state, the proposed neuron performs better in both metrics. It further improves over the previous LIF neuron demonstration[13] using high on-state current based impact ionization and above threshold circuit design by enabling sub-threshold BTBT based SOI device operation as well as sub-threshold control circuits. The proposed neuron provides 10× higher area efficiency compared to CMOS design with equivalent energy/spike. Alternatively, it has $10^4 \times$ energy efficiency at area-equivalent neuron technologies (e.g., Impact ionization or phase change based neurons). The proposed neuron in this work has several advantages; it operates completely asynchronously which is an essential characteristic of biomimetic systems. Additionally, the leakiness of the LIF neuron plays an important role in biology by introducing a time-dependent memory for neurons based on ion channel dynamics[28]. Further, this leakiness enables the neurons to process noisy signals in a noisy environment[29]. Ultimately, the use of mature PD-SOI technology makes it favorable for very large-scale implementation of SNNs.

Table 1. Benchmarking with the state-of-the-art

| Sr. No. | Authors | Neuron Model | Synaptic Input | Platform | Circuit Type | Tech. Node | Area $\mu m^2$ ($F^2$) | Energy/Spike (fJ) |
|---|---|---|---|---|---|---|---|---|
| 1 | Indiveri, G. et al.[15] | LIF | Current | CMOS | Analog-Digital | 0.35 um | 2573 ($21\times10^3$) | $9\times10^5$ |
| 2 | Wijekoon, J. H. B. et al.[16] | LIF | Current | CMOS | Analog | 0.35 um | 2800 ($23\times10^3$) | $8\times10^3$-$9\times10^3$ |
| 3 | Joubert, A. et al.[9] | LIF | Current | CMOS | Digital | 65 nm | 538 ($127\times10^3$) | $4.13\times10^4$ |
| 4 | Tuma, T. et al.[20] | IF | Voltage | Phase change + CMOS | Analog-Digital | 14 nm | 0.5 – 1 ($2551\times10^3$-$5102\times10^3$) | $3\times10^4$ |
| 5 | Dutta, S. et al.[13] | LIF | Voltage | SOI CMOS | Analog | 32 nm | 1.8 (1767) | $3.5\times10^4$ |
| 6 | Lashkare, S. et al.[21] | IF | Voltage | PCMO + CMOS | Analog-Digital | 32 nm | 3.16 (3086) | $1\times10^4$ |
| 7 | Han, J. W. et al.[14] | LIF | Current | Si (npn) | Analog | - | - (4) | 140 – $5.68\times10^3$ |
| 8 | Cruz-Albrecht, J. M. et al.[5] | LIF | Voltage | CMOS | Analog | 90 nm | 442 ($54\times10^3$) | 400 |
| 9 | Sourikopoulos, I. et al.[6] | LIF | Current | CMOS | Analog | 65 nm | 35 (8284) | 4 |
| 10 | **This Work** | **LIF** | **Current/Voltage** | **SOI CMOS** | **Analog** | **32 nm** | **0.8*** **(784)** | **3.22** |

* Projected area at 32 nm node

## Conclusion

To summarize, we have experimentally demonstrated a PDSOI MOSFET based LIF neuron on a highly manufacturable 32nm CMOS SOI technology. Novel use of subthreshold-based operation using band-to-band tunneling phenomenon at the device level as a well sub-threshold control-circuit operation enables extremely energy-efficient operation with 3.2 fJ/spike energy. The use of the compact circuitry along with extremely high energy efficiency makes this neuron an attractive choice for large-scale implementation of SNNs in hardware. This enables a 10× higher area efficiency compared to state-of-the-art CMOS design with equivalent energy/spike. Alternatively, $10^4$ × better energy efficiency is observed at area-equivalent neuron technologies.

## Methods

The 32nm Silicon-On-Insulator (SOI) High-K Metal Gate (HKMG) CMOS technology [30,31] is used to fabricate the devices used in this study. The gate dielectric stack consists of chemically grown 1.7nm HfO$_2$ (ALD) and 0.8nm interfacial SiO$_2$. The $V_T$ of the device is adjusted by using a Lanthanum capping layer between the HfO$_2$ gate oxide and the TiN metal gate. High volume manufacturability and excellent CMOS performance are demonstrated earlier [32]. The cross-sectional TEM image of the fabricated device (PDSOI MOSFET) is shown in Fig. 6.

All the DC-IV and Transient IV characterizations are performed at room temperature using the Agilent B1500A Semiconductor Device Parameter Analyser.

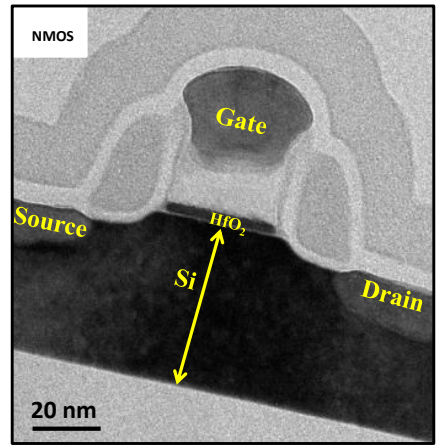

Fig. 6: TEM image of the fabricated device (PDSOI MOSFET) with gate oxide ($HfO_2$) thickness of 1.7 nm at 32 nm technology node [13].

## Acknowledgments

The authors wish to acknowledge the Science and Engineering Research Board, Department of Science and Technology and the Department of Electronics and IT, Government of India for providing a fund for this work.

## References

1. Maass, W. Networks of Spiking Neurons: The Third Generation of Neural Network Models. *Neural Networks* **10,** 1659–1671 (1997). doi: 10.1016/S0893-6080(97)00011-7

2. Herculano-Houzel, S. The human brain in numbers: a linearly scaled-up primate brain. *Front. Hum. Neurosci.* **3,** 1–11 (2009). doi:10.3389/neuro.09.031.2009

3. Merolla, P. *et al.* A digital neurosynaptic core using embedded crossbar memory with 45pJ per spike in 45nm. in *IEEE Custom Integrated Circuits Conference (CICC)* 1–4 (2011). doi:10.1109/CICC.2011.6055294

4. Shin, J. & Koch, C. Dynamic Range and Sensitivity Adaptation in a Silicon Spiking Neuron. *IEEE Trans. Neural Networks* **10,** 1232–1238 (1999). doi:10.1109/72.788662

5.  Cruz-Albrecht, J. M., Derosier, T. & Srinivasa, N. A scalable neural chip with synaptic electronics using CMOS integrated memristors. *Nanotechnology* **24,** (2013). doi:10.1088/0957-4484/24/38/384011

6.  Sourikopoulos, I. *et al.* A 4-fJ/spike artificial neuron in 65 nm CMOS technology. *Front. Neurosci.* **11,** (2017). doi:10.3389/fnins.2017.00123

7.  Hynna, K. M. & Boahen, K. Silicon neurons that burst when primed. *IEEE Int. Symp. Circuits Syst.* 3363–3366 (2007). doi:10.1109/ISCAS.2007.378288

8.  Hynna, K. M. & Boahen, K. Neuronal Ion-Channel Dynamics in Silicon. *IEEE Int. Symp. Circuits Syst.* 3614–3617 (2006). doi:10.1109/ISCAS.2006.1693409

9.  Joubert, A., Belhadj, B., Temam, O. & Heliot, R. Hardware spiking neurons design: Analog or digital? *IEEE Int. Jt. Conf. Neural Networks* 1–5 (2012). doi:10.1109/IJCNN.2012.6252600

10. Joubert, A., Belhadj, B. & Héliot, R. A robust and compact 65 nm LIF analog neuron for computational purposes. *IEEE Int. New Circuits Syst. Conf.* 9–12 (2011). doi:10.1109/NEWCAS.2011.5981206

11. Basu, A., Shuo, S., Zhou, H., Hiot Lim, M. & Huang, G. Bin. Silicon spiking neurons for hardware implementation of extreme learning machines. *Neurocomputing* **102,** 125–134 (2013). doi:10.1016/j.neucom.2012.01.042

12. Ostwal, V., Meshram, R., Rajendran, B. & Ganguly, U. An ultra-compact and low power neuron based on SOI platform. *Int. Symp. VLSI Technol. Syst. Appl. Proc.* 1–2 (2015). doi:10.1109/VLSI-TSA.2015.7117569

13. Dutta, S., Kumar, V., Shukla, A., Mohapatra, N. R. & Ganguly, U. Leaky Integrate and Fire Neuron by Charge-Discharge Dynamics in Floating-Body MOSFET. *Sci. Rep.* **7,** 1–9 (2017). doi:10.1038/s41598-017-07418-y

14. Han, J. W. & Meyyappan, M. Leaky Integrate-And-Fire Biristor Neuron. *IEEE Electron Device Lett.* **39,** 1457–1460 (2018). doi:10.1109/LED.2018.2856092

15. Indiveri, G., Chicca, E. & Douglas, R. A VLSI Array of Low-Power Spiking Neurons and Bistable Synapses With Spike-Timing Dependent Plasticity. *IEEE Trans. Neural Networks* **17,** 211–221 (2006). doi:10.1109/TNN.2005.860850

16. Wijekoon, J. H. B. & Dudek, P. Compact silicon neuron circuit with spiking and bursting behaviour. *Neural Networks* **21,** 524–534 (2008). doi:10.1109/IJCNN.2007.4371151

17. Moon, K. *et al.* ReRAM-based analog synapse and IMT neuron device for neuromorphic system. *Int. Symp. VLSI Technol. Syst. Appl.* 9–10 (2016). doi:10.1109/VLSI-TSA.2016.7480499

18. Lin, J. *et al.* Low-voltage artificial neuron using feedback engineered insulator-to-metal-transition devices. *IEEE Int. Electron Devices Meet.* 1–4 (2017). doi:10.1109/IEDM.2016.7838541

19. Pickett, M. D., Medeiros-Ribeiro, G. & Williams, R. S. A scalable neuristor built with Mott memristors. *Nat. Mater.* **12,** 114–117 (2013). doi:10.1038/nmat3510

20. Tuma, T., Pantazi, A., Le Gallo, M., Sebastian, A. & Eleftheriou, E. Stochastic phase-change neurons. *Nat. Nanotechnol.* **11,** 693-699 (2016). doi:10.1038/NNANO.2016.70

21. Lashkare, S. *et al.* PCMO RRAM for Integrate-and-Fire Neuron in Spiking Neural Networks. *IEEE Electron Device Lett.* **39,** 484–487 (2018). doi:10.1109/LED.2018.2805822

22. Chavan, T., Dutta, S., Mohapatra, N. R. & Ganguly, U. An ultra energy efficient neuron enabled by tunneling in sub-threshold regime on a highly manufacturable 32 nm SOI CMOS technology. *Device Res. Conf. - Conf. Dig. DRC* 1–2 (2018). doi:10.1109/DRC.2018.8442229

23. Maass, W. & Bishop, C. M. *Pulsed Neural Networks*. MIT Press, Massachusetts **275** (1999).

24. Mead, C. *Analog VLSI Implementation of Neural Systems*. Kluwer Academic Publishers (1989). doi:10.1007/978-1-4613-1639-8

25. Shukla, A., Kumar, V. & Ganguly, U. A software-equivalent SNN hardware using RRAM-array for asynchronous real-time learning. *IEEE Int. Jt. Conf. Neural Networks* 1–8 (2017). doi:10.1109/IJCNN.2017.7966447

26. Shukla, A., Prasad, S., Lashkare, S. & Ganguly, U. A case for multiple and parallel RRAMs as synaptic model for training SNNs. *Int. Jt. Conf. Neural Networks* 1–8 (2018). doi:10.1109/IJCNN.2018.8489429

27. Shukla, A. & Ganguly, U. An On-Chip Trainable and the Clock-Less Spiking Neural Network With 1R Memristive Synapses. *IEEE Transactions on Biomedical Circuits and Systems* (2018). doi:10.1109/TBCAS.2018.2831618

28. Hodgkin, A. L. & Huxley, A. F. A quantitative description of membrane current and its application to conduction and excitation in nerve. *J. Physiol.* **117,** 500–544 (1952). doi:10.1080/00062278.1939.10600645

29. Rauch, A. Neocortical Pyramidal Cells Respond as Integrate-and-Fire Neurons to In Vivo-Like Input Currents. *J. Neurophysiol.* **90,** 1598–1612 (2003). doi:10.1152/jn.00293.2003

30. Butt, N. *et al.* A 0.039um2 High Performance eDRAM Cell based on 32nm High-K/Metal SOI Technology. *IEEE Int. Electron Devices Meet.* 616–619 (2010). doi:10.1109/IEDM.2010.5703434

31. Greene, B. *et al.* High Performance 32nm SOI CMOS with High-k/Metal Gate and 0.149μm2 SRAM and Ultra Low-k Back End with Eleven Levels of Copper. *2009 Symp. VLSI Technol.* 140–141 (2009).

32. Horstmann, M. *et al.* Advanced SOI CMOS transistor technologies for high-performance microprocessor applications. *Proc. Cust. Integr. Circuits Conf.* **2,** 149–152 (2009). doi:10.1109/CICC.2009.5280865

**Author Contribution**

S.D. and T.C. conceived the idea. U.G. developed the idea. TC conducted the experiments, analyzed the results, performed the simulations, and wrote the manuscript taking help from N.M. and U.G. N.M. helped in fabrication. All the authors discussed the results and contributed to this work.

**Competing Interests:** The authors declare no competing interests.

# Supplementary Information

# Band-to-Band Tunneling based Ultra-Energy Efficient Silicon Neuron

Tanmay Chavan[1]*, Sangya Dutta[1], Nihar R. Mohapatra[2], and Udayan Ganguly[1]†

[1]Department of Electrical Engineering, IIT Bombay, Mumbai 400076, India
[2]Department of Electrical Engineering, IIT Gandhinagar, Gandhinagar 382355, India

*tanmay.chavan@iitb.ac.in; †udayan@ee.iitb.ac.in

# Supplementary Information 1:

A. Experimental Energy/Spike for Leaky-Integration

The effective neuron input current ($I_{in}$; tunneling current) that charges the body of the proposed device (integrator) is extracted by sampling the experimental DC drain current ($I_D$) vs. source voltage ($V_S$) data (with gate grounded) at $V_S$ =0.4 V (Bias point for neuron operation) as shown in Figure S1. (a). The effective neuron input current ($I_{in}$) as a function of drain bias ($V_D$) is given in Figure S1. (b). The energy/spike for the integrator (Figure S1. (c)) is calculated by integrating the power supplied for integration ($V_D \times I_{in}$) over the duration of one spike cycle ($t_{cycle}$). ($t_{cycle} = \frac{1}{Spike\ Frequency}$)

Average Energy/Spike for leaky-integration= 2.74 fJ ($V_D = 0.8\ V\ to\ V_D = 1.5\ V$)

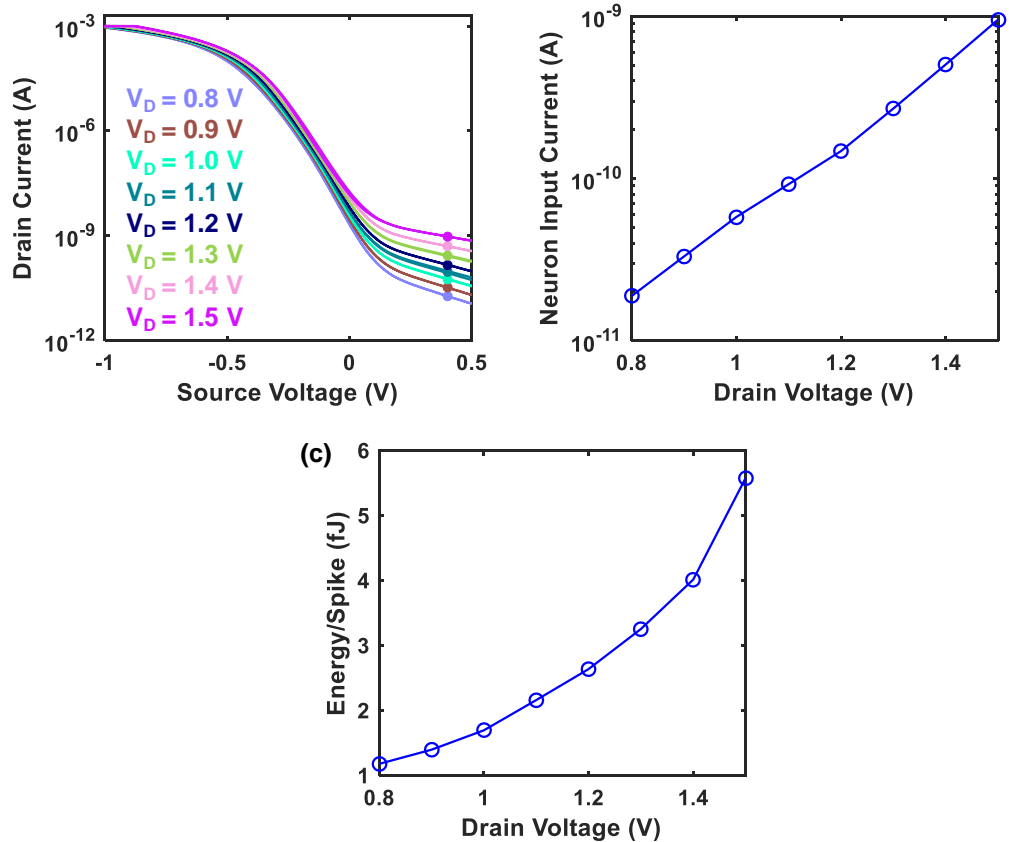

Fig. S1. (a) Experimental DC Drain current vs. Source voltage characteristics for different drain bias within neuron operating region (b) Sampled drain current i.e. neuron input current is obtained by sampling the curves in (a) at source voltage = 0.4 V, (c) Energy/Spike consumed by the integrator obtained by integrating the supplied power over duration of one spike cycle.

B. Circuit Realization of BTBT Neuron

Figure S2. (a) shows the circuit diagram for the Leaky-Integrate and Fire (LIF) neuron with the current based operation. The circuit implementation for the threshold detector is given in Figure S2. (b). The input current ($I_{input}$) is the net synaptic current from the various input synapses. The neuron uses the proposed SOI device (M1) in a configuration where the input current is applied at the drain terminal, membrane potential is read out from the body terminal, and the gate-source bias is used to operate the device in the subthreshold region. The use of subthreshold design and tunneling phenomenon enables extremely energy efficient operation. The neuron circuit comprises the following blocks: (i) *Integrator*, (ii) *Threshold Detector*, (iii) *Reset & Leakiness*. The contribution of each block towards LIF neuron operation is given below.

- Leaky-Integration

The equivalent circuit in the leaky-integration phase is shown in Figure S2. (c). When the input current is applied at the drain terminal of the device (M1), the floating drain voltage adjusts itself to balance the incoming input current with the source-drain current of M1. For large input currents, the drain voltage charges up to a large value which causes significant band-to-band tunneling between the drain-body region. This current leads to the accumulation of holes in the body and charges the capacitor $C_B$ (*integration*). A small fraction of the stored holes leaks out through the leak resistance of MOSFET M2 (*leakiness*).

- Fire & Reset

The equivalent circuit for fire & reset phase is shown in Figure S2. (d). The charging of capacitor $C_B$ causes the body potential to increase. A cascade of two inverters (INV1 and INV2) is used for firing threshold ($V_{th}$) detection. When the body potential exceeds the switching threshold ($= V_{th}$) of INV1, the output of INV2 goes high (*fire*). This drives the gate of M2 which leaks out the holes stored in the body of M1 (*reset*).

The circuit in Figure S2. (a) is implemented in Cadence Virtuoso using the GlobalFoundries 180 nm GFUS7SW technology. Transient simulation is performed to validate the LIF neuron functionality proposed in this work. Also, the energy- and area – efficiency of the neuron is calculated using the simulation results. The device parameters used for circuit simulation are tabulated in Table S1. The transient dynamics for the body potential ($V_{body}$) and the corresponding spiking output ($V_{spike}$) for an input current of 1 nA is shown in Figure S3. The area for the circuit is estimated from the layout shown in Figure S4.

Table S1: MOSFET parameters for circuit simulation

| Name | L ($\mu m$) | W ($\mu m$) | Type |
|---|---|---|---|
| M1 | 0.32 | 0.5 | NMOS |
| M2 | 0.4 | 0.5 | NMOS |
| M3 | 0.4 | 0.5 | NMOS |
| M4 | 0.4 | 0.5 | NMOS |
| M5 | 0.4 | 0.5 | PMOS |
| M6 | 0.32 | 0.5 | PMOS |

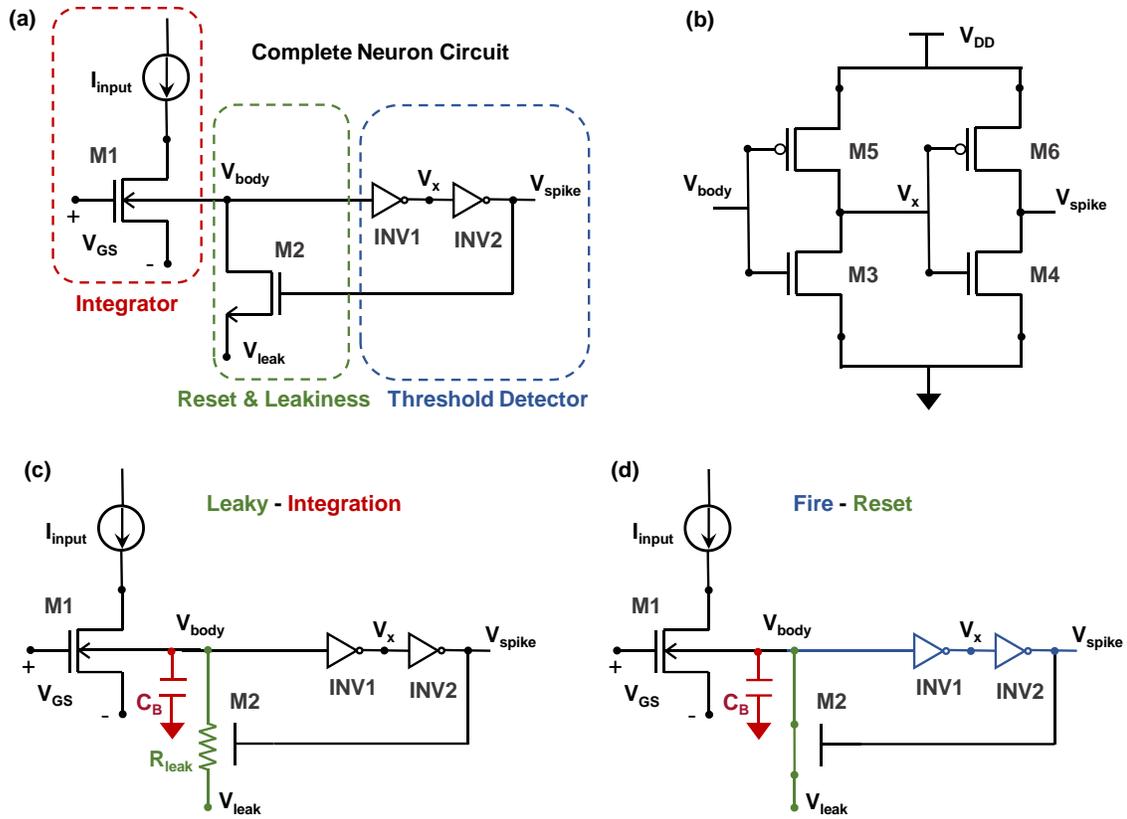

Fig. S2. (a) Complete neuron circuit consisting of the integrator, reset & leakiness, and threshold detector block ($V_{GS} = 0\ V; V_{leak} = -0.125\ V$), (b) circuit implementation of INV1 and INV2 ($V_{DD} = 0.25\ V$), (c) Equivalent circuit for leaky-integration phase, (d) Equivalent circuit for fire-reset phase.

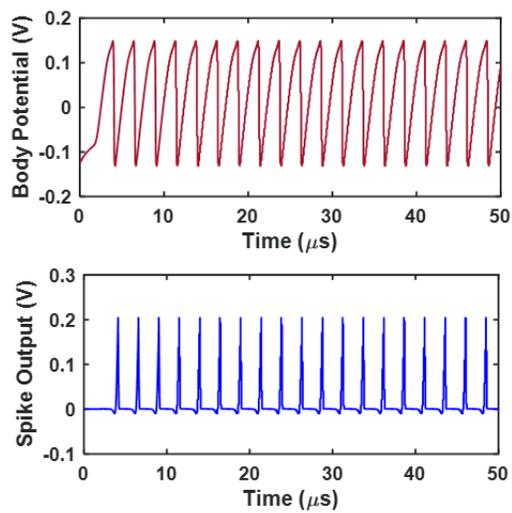

Fig. S3. Spectre$^{TM}$ simulation results for circuit in Fig. S2. (a) showing leaky-integration and reset in body potential ($V_{body}$) and corresponding spiking output ($V_{spike}$).

Average Energy/Spike for Leaky integration (SOI Device) = 2.74 fJ

Energy/Spike for Reset and Spike (Reset Circuit) = 0.48 fJ

Total Average Energy/Spike = 3.22 fJ

- Layout and Area Estimate for BTBT Neuron

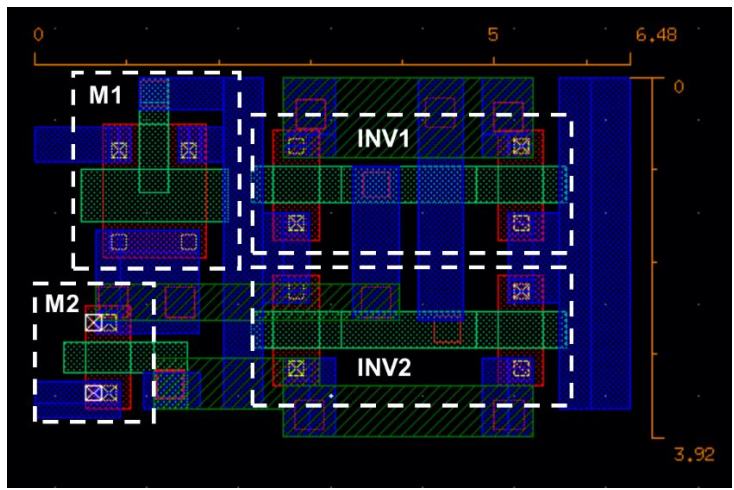

Fig. S4. Layout for the neuron circuit in Fig. S2. (a)

- o The area at 180nm node: $6.48 \mu m \times 3.92 \mu m = 25.4016 \ \mu m^2$
- o Normalized area $= \frac{25.4016 \ \mu m^2}{0.18 \mu m \times 0.18 \mu m} = 784 \ F^2$

# Supplementary Information 2

## Performance Evaluation of BTBT Neuron in a Spiking Neural Network (SNN):

- LIF Model for SNN Simulations

The behavioral neuron model for SNN simulation is shown in Figure S5. (a). The drain dependent tunneling current is modeled by a voltage-dependent current source ($I_{in} = f(V_D)$). The input current is applied to a parallel combination of a leak resistor ($R_L$) and membrane capacitance ($C_M$). The resting potential for the neuron is set by the DC source ($E_L$). Once the membrane potential (measured across $C_M$) exceeds the firing threshold ($V_{th}$) the neuron is reset by leaking out the charge stored in $C_M$ through the switch SW[1].

The input current is obtained experimentally from the DC characteristics as shown in Figure S1. (a) and (b). The experimental spike frequency of the neuron (Fig. 6(e) in the manuscript) is modeled by the equation Eqn. S1 derived for the circuit in Figure S5. (a). The spiking response of the model matches with the experimental data as shown in Figure S5. (b).

$$f = \frac{1}{R_P + \tau_M \times \ln\left(\frac{E_L - I_{in}R_L}{V_{th} - I_{in}R_L}\right)} \qquad (\text{Eqn. S1})$$

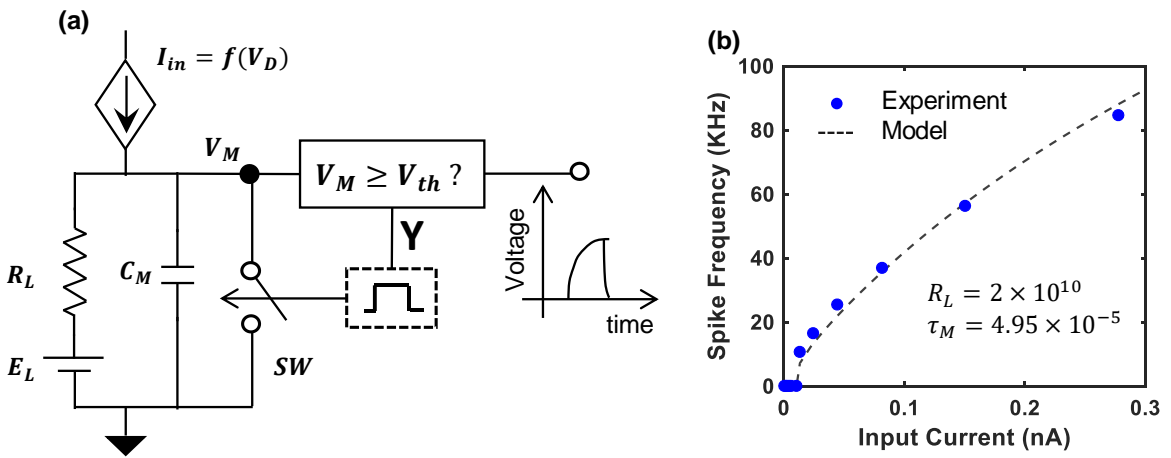

Fig. S5. (a) Behavioral circuit model for LIF neuron simulation in software, (b) response of the model (dashed line) fits with the experimental data (closed blue dots)

Where $\tau_M = R_L \times C_M$

- SNN Architecture and Simulation Details

A two-layer feedforward (16×3) SNN is implemented in MATLAB to solve Fischer's Iris Classification problem[1]. The Fischer's Iris data set consists of 3 classes of flower (50 samples each). Each sample is characterized by 4 features, Petal Length, Petal Width, Sepal Length, Sepal Width as shown in Figure S6. (a). The raw features are the first population coded by normalization and transformation. The transformation step converts the raw data into input currents for the first layer of the SNN. The 16 input neurons are connected to 3 output neurons through the excitatory synapses as shown in Figure S6. (b). The output layer forms a winner take all configuration through mutual inhibition. Learning takes place through a supervised Spike-Time-Dependent-Plasticity rule. Additional information about population coding is given in Figure S6. (c). The learning accuracy as a function of training iterations is plotted in Figure S7. 96% accuracy is achieved for the Fisher Iris classification dataset, which matches state of the art for SNN[2,3].

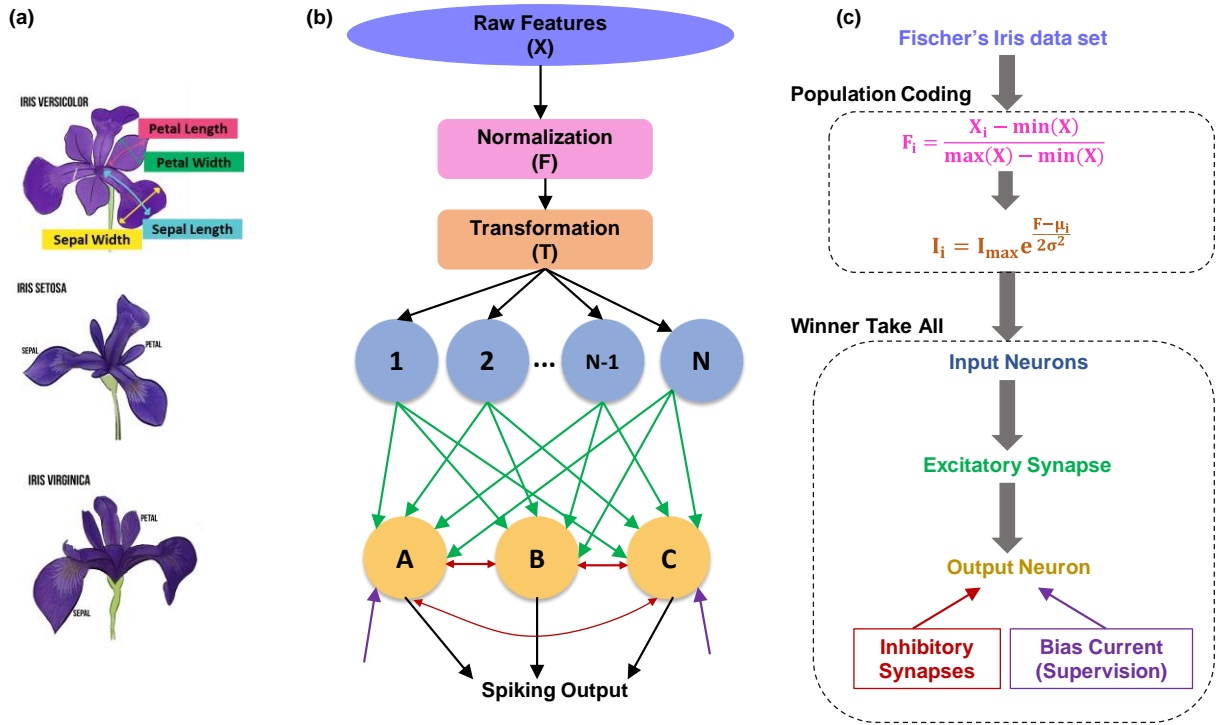

Fig. S6. (a) Three classes of flowers in Fischer's Iris data set with features of each flower, (b) Network architecture for the SNN to be simulated in MATLAB, (c) Algorithmic details to SNN architecture.

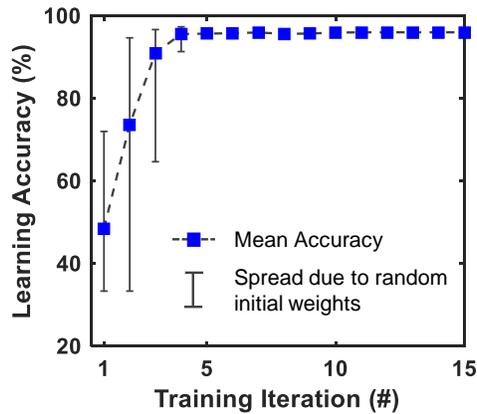

Figure S7. Increasing learning accuracy for recognition task on Fisher's Iris data set with increasing training iterations. Maximum learning accuracy of 96% is achieved with the proposed neuron model in MATLAB.

## References

1. Koch, C., Segev, I., (1999) - Methods in neuronal modeling; from ions to networks (2nd ed.). Cambridge, Massachusetts: MIT Press. p. 687. ISBN 0-262-11231-0

2. Biswas, A., Prasad, S., Lashkare, S. & Ganguly, U. A simple and efficient SNN and its performance & robustness evaluation method to enable hardware implementation. (**arXiv:1612.02233v1** **[cs.NE]**)

3. Xin, J. & Embrechts, M. J. Supervised learning with spiking neural networks. Int. Jt. Conf. Neural Networks. Proc. (Cat. No.01CH37222) **3**, 1772–1777 (2001).